\documentclass[preprint,12pt]{elsarticle}

\pdfoutput=1

\usepackage{amssymb}

\usepackage{multirow}
\usepackage{subfigure}
\PassOptionsToPackage{hyphens}{url}
\usepackage{hyperref}

\journal{Elsevier Journal}

\title{Two-Stage Classifier for COVID-19 Misinformation Detection Using BERT: a Study on Indonesian Tweets}

\author[]{Douglas Raevan Faisal}
\author[]{Rahmad Mahendra}

\begin{document}

\begin{frontmatter}



\begin{abstract}
The COVID-19 pandemic has caused globally significant impacts since the beginning of 2020. This brought a lot of confusion to society, especially due to the spread of misinformation through social media. Although there were already several studies related to the detection of misinformation in social media data, most studies focused on the English dataset. Research on COVID-19 misinformation detection in Indonesia is still scarce. Therefore, through this research, we collect and annotate datasets for Indonesian and build prediction models for detecting COVID-19 misinformation by considering the tweet’s relevance.
The dataset construction is carried out by a team of annotators who labeled the relevance and misinformation of the tweet data. 
In this study, we propose the two stage classifier model using IndoBERT pre-trained language model for the Tweet misinformation detection task. We also experiment with several other baseline models for text classification. The experimental results show that the combination of the BERT sequence classifier for relevance prediction and Bi-LSTM for misinformation detection outperformed other machine learning models with an accuracy of 87.02\%. 
Overall, the BERT utilization contributes to the higher performance of most prediction models. We release a high-quality COVID-19 misinformation Tweet corpus in the Indonesian language, indicated by the high inter-annotator agreement.
\end{abstract}




\end{frontmatter}


\section{Introduction}
\label{sec:introduction}

The global effect of the COVID-19 pandemic had been one of the most significant events starting early 2020s. This also applies even more severely in Indonesia, with over 4 million confirmed cases nationally at the end of 2021 \cite{kompas2021update}. The pandemic situation that has been on for almost 2 years resulted in more discoveries being made as it goes on. Those discoveries are ranging from medical breakthroughs such as vaccines \cite{who2021vaccines}, to the mutations that introduced new coronavirus variants \cite{who2021variants}.

While new information and findings get spread widely, other false information or misinformation is also spread across mediums. In Indonesia, social media has been one of the technologies that have the fastest adoption rate and also became the easiest medium for misinformation to propagate quickly \cite{juditha2018interaksi}. However, on the other hand, it is found that as much as 76\% of their survey respondents believed social media was a trusted information source \cite{literasiDigital2020}. The pandemic also affected their findings in which the health and medical category became the second most prominent issue after politics in the spread of misinformation in Indonesia \cite{literasiDigital2020}. As the technology became more accessible, there was a tendency for people to look for information through social media as opposed to other traditional media \cite{jacobs2017health}.

Detecting misinformation in COVID-19 topics using machine learning approaches has been examined by several research works, specifically in the English social media text corpus.
Using the social media post features without any prior knowledge of the facts apart from the dataset itself, machine learning models could achieve above 90\% scores in accuracy \cite{al2020lies,elhadad2020detecting}.
The use of pre-trained language models, i.e., Bidirectional Encoder Representations from Transformers (BERT) \cite{devlin2018bert} improved the misinformation classification performance when compared to predecessor methods, such as word embeddings and $n$-grams models \cite{song2021classification}.

Following the COVID-19 pandemic, Panacea Lab worked on collecting COVID-19 social media posts on Twitter \cite{banda2020large} and updated their dataset periodically \footnote{\url{http://www.panacealab.org/covid19/}}. 
As of December 2021, the Indonesian language was the sixth most written language among 65 distinct languages in their dataset. Other work inspected that Indonesian was the third most popular language in their COVID-19 Twitter dataset from January to May 2020 \cite{chen2020tracking}. Their findings indicated that the Indonesian language has a large proportion of the public discourse in social media.

This research fills the gap in detecting misinformation in Indonesian social media Twitter posts by using the BERT language model. There are two main contributions of this research:
\begin{enumerate}
\item Dataset building. We first collected Indonesian social media text data related to the COVID-19 pandemic from Twitter. We annotated the sample of tweets for the misinformation detection task. The data collection spanned more than one year of Indonesian COVID-19-related tweets, and the annotation process follows a comprehensive annotation guideline. The Indonesian COVID-19 Twitter corpus\footnote{\url{https://github.com/douglasraevan/covid19-indonesian-tweets}} and misinformation detection dataset\footnote{\url{https://github.com/douglasraevan/covid19-indonesian-misinformation-tweets}} are publicly available for research purpose. 
\item Model building. This research revolves around detecting misinformation with the machine learning approach. The text mining experiments in this research present insights into how well a machine learning model combined with a pre-trained language model can learn and detect misinformation in social media. 
\end{enumerate}

\section{Related Work}

Misinformation is defined as a false or inaccurate information \cite{karlova2013social, kumar2014detecting}. In the science and health context, misinformation is defined as information that contradicts the consensus agreement of scientific communities on a phenomenon \cite{swire2020public}. Other related terms are hoax and disinformation. Disinformation is intentionally made to be deceptive, while misinformation may or may not be deceptive. Karlova and Fisher argued that misinformation and disinformation should be viewed as a separate set \cite{karlova2013social}, whereas Kumar and Geethakumari \cite{kumar2014detecting} and Swire-Thompson and Lazer \cite{swire2020public} considered disinformation as the subset of misinformation. Due to the difficulty of the current computational model in determining someone's deceptive intention in text, most automatic misinformation detection researches lean towards the latter definition and ignored the deceptive factors of disinformation.


There are four mediums in which misinformation can propagate, which are direct to online sources, search engines, user-generated content, and mobile apps \cite{swire2020public}. A study found that misinformation was spreading faster than the true information by tracking around 126,000 online rumors on Twitter between 2006 and 2017 \cite{vosoughi2018spread}.


As a platform that played a big role in user-generated content, social media is defined as the use of electronic and Internet tools to share and discuss information and experiences with other people \cite{moturu2009quantifying}. Among the variety of data types shared in social media, the text is one of the most used data types 
and therefore can be seen as a fundamental part of most platforms \cite{hu2012text}. The large amount of text data stored in social media became the main motive for the research in social media analytics through text mining. Following the traditional framework of text analytics, \cite{hu2012text} defined the social media text analytics framework consisted of three consecutive phases, which are text preprocessing, text representation, and knowledge discovery.

\subsection{Data Annotation}

Machine learning relies heavily on data samples to be able to create meaningful inferences. Annotating data for the misinformation classification task is challenging. Full Fact and academic institution in the UK designed a guide for crowdsourcing data annotation on verifiable factual claims \cite{konstantinovskiy2020automated}. Each claim is categorized into seven different classes, which are "personal experience", "quantity in the past or present", "correlation or causation", "current laws or rules of operation", "prediction", "other type of claim", and "not a claim". 

QCRI team proposed several improvements to the annotation process of Tweet misinformation data \cite{alam2021fighting}. They adjusted the guideline and asked seven questions to annotators when annotating the Tweet data. The main idea is to guide the annotators to determine whether a tweet contains any verifiable factual claim, then proceed to verify whether the claim is true or not and estimate the significance of the impact if it were to be seen by the majority of social media users. One thing that differentiates the QCRI's annotation guideline from Full Fact`s is the exclusion of "personal experience" from being labeled as misinformation. The argument is that personal experiences mostly consist of claims about facts that could not be verified with any publicly-available information.

\subsection{Detecting Misinformation}

Detecting misinformation on social media data has been done by several works on English text corpus. Al-Rakhami and Al-Amri collected around 409,484 tweets that revolved around the COVID-19 topic and constructed an ensemble-based model to classify whether they contain any misinformation \cite{al2020lies}. They only used structured data as the features that became the input for the ensemble model. Another work \cite{aldwairi2018detecting} performed a similar method in classifying fake news based on the clickbait elements that are contained within the news titles and URLs. They extracted shape features of the text (e.g. uppercase, lowercase, numeric) and keywords matching as the features for the machine learning.

Other work explored the use of textual data as the main feature for the machine learning algorithm. The Indian researchers found that SVM model using TF-IDF representation yields satisfying accuracy \cite{sharif2021combating} when being tested on the annotated fake news corpus built from several established fact-checking websites \cite{patwa2021fighting}.

More modern approaches explored the use of word embeddings and pre-trained language models. As current state-of-the-art language models, the Transformer \cite{vaswani2017attention} and BERT \cite{devlin2018bert} have been highly referenced and used by multiple NLP downstream tasks. Several works have utilized BERT for misinformation detection \cite{hossain2020covidlies,song2021classification}. Song constructed classification-aware neural topic model which hybridizes a deep-learning-based classifier and topic modeling \cite{song2021classification}. The model was then trained to classify 10 categories of misinformation and then compared to BERT as the existing state-of-the-art model.


COVIDLies was introduced as the dataset for stance detection task in which the model is challenged to identify whether a tweet support, refute, or has no stance towards retrieved misconceptions about COVID-19 \cite{hossain2020covidlies}. The BERT model was utilized with domain adaptation to improve the accuracy performance for this domain specific task. Another work collected particular Bulgarian news articles and then classified them into credible or fake news \cite{hardalov2016search}. They experimented with three types of features, i.e., linguistic feature (TF-IDF), credibility features (orthographic), and semantic features (Word2Vec). 



Two key points that were overlooked by the most previous works are the imbalanced ratio nature between true information compared to misinformation and lack of topic relevance judgement in misinformation dataset. Naturally, the number of misinformation is less significant than the number of available true information. It has implication that the misinformation detection should be considered as an imbalanced classification problem. On the other hand, misinformation detection should not be viewed only as a binary classification task, but rather as two binary classification tasks. The first classification task is to determine whether the data is relevant enough to be verified. If it fulfills the criteria of being relevant, then the second task is to verify and classify whether the claim is true to the fact or false.


\subsection{Bidirectional Encoder Representations from Transformers}

Bidirectional Encoder Representations from Transformers (BERT) is a language model or text representation model that learn the context and semantic meaning of documents bidirectionally \cite{devlin2018bert}. This model utilizes the self-attention layers architecture from the Vaswani Transformer model \cite{vaswani2017attention}. In addition, BERT has 
bidirectional architecture for fine-tuning multiple tasks that require context from two directions such as question answering. This model also implements language representation with a pre-training approach for reuse in several different language processing tasks. Similar to the Word2Vec model, the output representation of this model is a numeric vector for each token in the document.

The initial phase involves a pre-training process by performing two unsupervised tasks, namely the Masked Language Model (MLM) and Next Sentence Prediction (NSP). The next phase is the fine-tuning process to adjust to the language processing task.

For Indonesian language, the IndoBERT model \cite{wilie2020indonlu} is trained with a dataset of Indonesian language texts, which includes social media texts, blogs, news articles, and online sites. The IndoBERT model outperforms several baseline models which includes fastText and mBERT in several NLP tasks in Indonesian language.

\section{Methodology}

\begin{figure}
    \centering
    \includegraphics[width=0.8\textwidth]{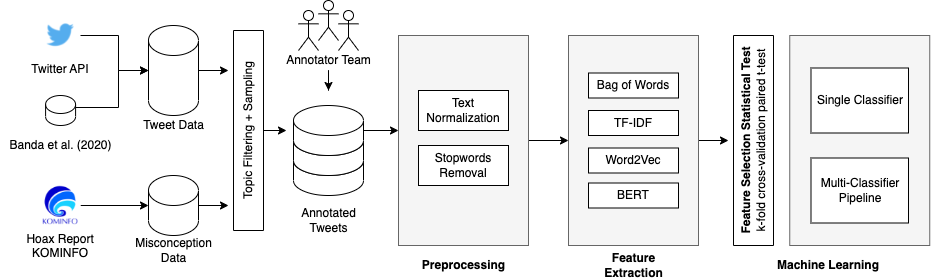}
    \caption{Overall View of The Research Process.}
    \label{figMethod}
\end{figure}

The design of this research process, as shown in Figure \ref{figMethod}, followed the framework for social media text mining by \cite{hu2012text}. The process consists of data collection, data annotation, preprocessing, feature extraction, and machine learning. The main contribution of this research is twofold, i.e., Indonesian COVID-19 misinformation dataset and machine learning model to detect misinformation. In addition, we also release large scale COVID-19 Indonesian Tweets collection.

We formulate the misinformation detection task as a text classification task. Each tweet is seen as a document instance containing a sequence of words. For each tweet, the input is a sequence of $N$ tokens $W = (w_1, w_2, ..., w_N)$ and the output is a single label prediction $y$ where $y \in \{$irrelevant, true, misinformation$\}$.  

\subsection{Data Collection}

\subsubsection{Tweet Data Retrieval}

As the primary dataset for this research, tweet data were collected directly through Twitter API similar to previous works \cite{banda2020large, chen2020tracking, qazi2020geocov19}. The data collection was conducted from July 21, 2020 to February 7, 2021. Due to the limitations of the Twitter API's free tier, we did not manage to retrieve tweets before July 21, 2020. 
To retrieve the Tweets collection, we defined the keywords that are related to the COVID-19 pandemic. We first started with several keywords as query seed and then we updated the keywords iteratively by evaluating the top n-grams periodically during the collection process.

We only considered tweets that were identified as written in the Indonesian language based on Twitter's language identification algorithm. However, Twitter language tags are not always accurate \cite{Zubiaga2016TweetLIDAB}, especially for closely related language pairs, e.g. Indonesian and Malay
\cite{Ranaivo-Malancon_2016}. We found that several tweets were actually written in the Malaysian language detected as Indonesian. To exclude those tweets from our collection, we identified tweets containing keywords that were specific to the Malaysian context, such as "malaysia", "KKMPutrajaya" (the Twitter handle for the Malaysian Ministry of Health), and "\textit{kes baharu}" (meaning: new case(s)). 

To complement the missing data before July 2020, we used the Indonesian portion dataset published by Panacea Lab \cite{banda2020large}. It covered COVID-19 tweets in multiple languages from January 27, 2020. The dataset has metadata of tweet IDs, timestamps, geolocation, and language, so a hydration process was required to collect the respective tweets based on the given tweet IDs.

Furthermore, we filtered the Panacea Lab data using the query we used to collect the data from Twitter API. We merged filtered Panacea Lab data with crawled data and obtain a total of 13,274,237 tweets. 


\subsubsection{Misconception Retrieval}

We followed COVIDLies \cite{hossain2020covidlies} methodology to harness a set of refuted false claims or misconceptions of COVID-19 in order to limit only related discussion topics to be included in the misinformation experiment. We use the misconceptions topics published by Kominfo, the Indonesian Ministry of Communication and Information Technology, during the timeframe of our Tweet collection (from January 27, 2020 to February 7, 2021)\footnote{\url{https://www.kominfo.go.id/content/all/laporan\_isu\_hoaks}}.

In addition, we used the misconceptions from COVIDLies for the global context. The data that they retrieved were originally published on a Wikipedia page on COVID-19 misinformation. We translated the misconceptions into Indonesian using the Google Translate API and then incorporated them into the Indonesian local misconception topics. Other COVID-19 misconceptions from the medical science perspective were also retrieved \cite{nsoesie2020covid} and \cite{nasir2020misinformation}, e.g., "5g", "hydroxychloroquine", "mosquito", and "garlic".

We sampled some of the misconceptions and then manually defined the list of keywords related to that misconception. The keywords are then used to filtered the Tweets to be included in our misinformation dataset.  

\subsection{Data Annotation}

The data annotation involve two steps, namely relevance annotation and misinformation annotation. The first step was to annotate each tweet whether it is relevant to be included in the misinformation classification or not. Once the tweet had been annotated as relevant, then the tweet would be included in the second step, which was misinformation annotation. The final annotated data is also published on Github\footnote{\url{https://github.com/douglasraevan/covid19-indonesian-misinformation-tweets}}.
 
\subsubsection{Relevance Annotation}

For this part, 4,500 tweets were randomly sampled from the tweets that matched the misconception-based query. Two annotators were recruited to annotate the relevance aspect of each tweet. Both annotated the same set of tweets. The annotators were given the tweet hyperlink, the posting time and date, the full text, and the URL contained in each tweet. In order to reduce any human bias, the annotators should not consider any belief inferred from the Twitter profile user when doing annotation.

The relevance annotation was done by asking the annotator three questions towards the Tweet, i.e., whether it is written in Indonesian, whether it is topically relevant, and whether it contains verifiable factual claim \cite{alam2021fighting, konstantinovskiy2020automated}. 


\section{Experimental Results}
\label{sec:experimentalResults}

\subsection{Annotation Results}

The annotation process consisted of two steps; the first step was to annotate the relevance of the tweet and the second step was to annotate whether the claim within the tweet was misinformation or not. For the relevance annotation, two annotators annotated 4,500 tweets and obtained a Kappa agreement score of 0.8840. This means the annotators had a high level of agreement in annotating the relevance of the tweet \cite{mchugh2012interrater}. 
In the misinformation annotation, two annotators verified the truth of claims for 2,441 Tweets with relevant label. The Kappa score of the two annotators was 0.8467, which can be interpreted as strong inter-annotator agreement.

The tweet count and the label distributions based on the annotation are presented in Table \ref{annotationDistTable}. The results show that there is a notable imbalance between claims that were true and misinformations in which the "true" label appeared in the annotations almost four times more than the "misinformation" label. There were also 237 tweets in which the majority of the annotators within a group voted as "not sure" and 22 tweets where the annotators thought that the tweets needed expert annotations. Lastly, there were 3.24\% out of all the annotated tweets where the votes of the annotators could not reach a consensus. 

\begin{figure}
    \centering
    \includegraphics[width=\textwidth]{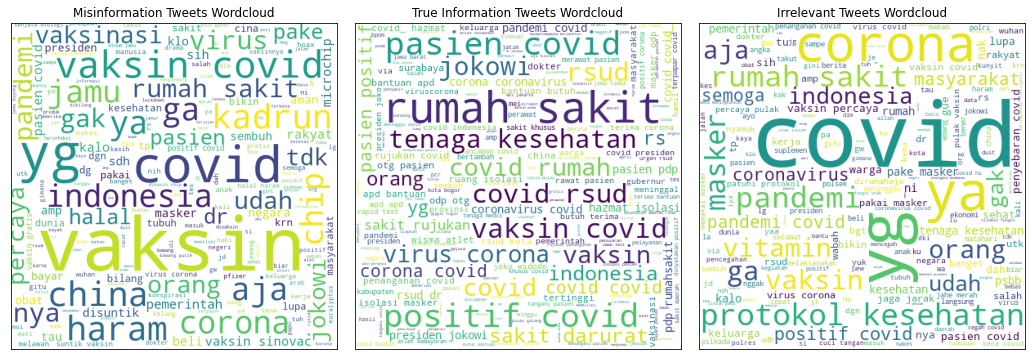}
    \caption{Annotated Tweets Wordcloud.}
    \label{tweetsWordcloud}
\end{figure}

We found several keywords or phrases that indicate different topics discussed for each of the labels as shown in the word clouds in Figure \ref{tweetsWordcloud}. Tweets that were annotated as misinformations were mostly discussing the vaccinations (keyword(s): "\textit{vaksin}", "\textit{vaksinasi}"), which at the time was still filled with rumors. Meanwhile, the tweets that were annotated as containing true information mostly discussed the recovery aspects from the pandemic, such as hospitals (\textit{rumah sakit} or \textit{rsud}), health workers (\textit{tenaga kesehatan}), and COVID patients (\textit{pasien covid}). As for the tweets that were labeled as irrelevant, they were mostly tweets on health protocols reminders, indicated by the dominance of keywords such as health protocol (\textit{protokol kesehatan}), face masks (\textit{masker}), and vitamin(s). 

\begin{table}[t]
\centering
\caption{Annotation Label Distribution}
\label{annotationDistTable}
\begin{tabular}{lrr}
Annotation Label      & Tweet Count & Percentage \\ \hline
Irrelevant            & 2,059       & 45.76\%    \\
True Information      & 1,632       & 35.60\%    \\
Misinformation        & 404         & 8.98\%     \\
Not Sure              & 237         & 5.93\%     \\
No Consensus          & 146         & 3.24\%     \\
Need Expert Judgement & 22          & 0.49\%    
\end{tabular}
\end{table}


We only include the Tweets that were labeled as irrelevant, true, and misinformation in final dataset and used it for the experiments. We did not perform further adjudication process for the remaining labels because the label counts for the other labels were not significant when compared to the main labels. We also further sampled several tweets which were already filtered out during the misconception filtering and then labeled the new samples as irrelevant. We aimed for the irrelevant label count to be around twice as much compared to the true label count. The final dataset contains 3,127 irrelevant Tweets. The dataset was splitted into three sets; train, validation, and test set with each label count shown in Table \ref{datasetSplit}. We also splitted them stratifiedly based on the label count, so each split has a near-identical label distribution.


\begin{table}[t]
\centering
\caption{Dataset Split}
\label{datasetSplit}
\begin{tabular}{lrrr}
             & \multicolumn{1}{l}{Irrelevant} & \multicolumn{1}{l}{True} & \multicolumn{1}{l}{Misinformation} \\ \hline
Train (60\%) & 1,876                           & 979                       & 242                                 \\ 
Val (20\%)   & 625                             & 327                       & 81                                  \\ 
Test (20\%)  & 626                             & 326                       & 81                                  \\ 
\end{tabular}
\end{table}


\subsection{Feature Selection Results}

Using the constructed dataset, we then experimented with some of the feature options combined with the classifiers. We tested the performance based on the F1 scores on classifying the validation set. The models were run using $k$-fold cross-validation and then prepared a pair of samples based on the features that we plan to compare. We assumed the null hypothesis ($H_0$) to denote the equal mean between the two sets and rejected the $H_0$ if the $P$-value produced by the statistical significance test is below 0.05.

The results are shown on Table \ref{statTest}. The first comparison was to evaluate the effect of implementing a balancing technique into the training dataset. Both SMOTE and random undersampling were used in a chain as the balancing technique. Based on the $t$-test results, we found that there was a significant effect by balancing the label distribution. Hence, $H_0$ was rejected. The mean of the balanced sample set was also found to be higher than the mean of the non-balanced sample set. Therefore, we determined to implement the balancing technique for the traditional features and classifiers.

The second comparison was between the Bag-of-Words (BoW) and TF-IDF. For the BoW, we implemented the binary type as opposed to the token occurrence count. We found that there was no significant difference in performance between BoW and TF-IDF. However, we investigated the results further and it turns out that the classifier models that were used in each feature type seemed to affect the scores. So the tests were then divided based on the classifiers. One group was specific to the Naive Bayes classifier, and the other group was all the other traditional classifiers. Both groups' results indicated that there were significant differences in different features (BoW and TF-IDF). It was found that the Naive Bayes classifier results were improved by using BoW, while the other classifiers results were improved by using TF-IDF.

\begin{table}[t]
    \centering
	\caption{Feature Selection Statistical Significance Test Results}
	\label{statTest}
	\begin{tabular}{llllll}
	    \multicolumn{2}{c}{Sample Sets} & $\bar{x}_1$ & $\bar{x}_2$ & $t$-value & $P$-value \\
		\hline
		Non-Balanced & Balanced & $22.29$ & $36.25$ & $8.8448$ & \textbf{0.0000*}\\
		BoW & TF-IDF & \multicolumn{4}{c}{}\\
	    \multicolumn{2}{l}{- All Classifiers} & $35.56$ & $33.94$ & $1.2487$ & $0.2129$\\
	    \multicolumn{2}{l}{- Naive Bayes} & $44.49$ & $25.70$ & $7.8314$ & \textbf{0.0000*}\\
	    \multicolumn{2}{l}{- SVM, LR, DT, RF, XGB} & $31.99$ & $37.23$ & $4.1978$ & \textbf{0.0000*}\\
	    \hline
        \multicolumn{6}{l}{\footnotesize{$\bar{x}_i$: The mean of F1 scores from sample set $i$}} \\
	    \multicolumn{6}{l}{\footnotesize{\textbf{*}: $P$-value $< 0.05$}} 
	\end{tabular}
\end{table}


\subsection{Single Classifier Results}


We first experimented with only a single classifier as our baseline to perform a 3-class classification task. The results were also divided into traditional machine learning classifiers and deep learning classifiers. Based on the results in Table \ref{singleClassifierResults}, the BERT pre-trained by \cite{wilie2020indonlu} achieved the highest accuracy score, misinformation recall, and misinformation F1 score. The XGBoost classifier with TF-IDF as its feature had the highest precision but with a high recall tradeoff. The opposite occurred with the SVM-Word2Vec combination where it had a high recall (the same as BERT) but with a low precision score. 

Most of the deep learning classifiers resulted a higher score when compared to the traditional classifiers. However, there were still several traditional classifiers that outperformed the deep learning classifiers. Based on the F1 scores, the best model for the traditional machine learning was the XGBoost model combined with the Word2Vec feature with an F1 score of 49.41\%, which was still lower than BERT, but higher than the other deep learning models. 


\begin{table}[t!]
\centering
\caption{Single Classifier Results}
	\label{singleClassifierResults}
\begin{tabular}{llllll}
\multicolumn{1}{l}{\multirow{2}{*}{Classifier}} & \multicolumn{1}{l}{\multirow{2}{*}{Feature}} & \multicolumn{1}{l}{\multirow{2}{*}{Acc.}} & \multicolumn{3}{l}{Target = Misinformation}                                                        \\ \cline{4-6} 
\multicolumn{1}{l}{}                            & \multicolumn{1}{l}{}                         & \multicolumn{1}{l}{}                     & \multicolumn{1}{l}{Prec.}           & \multicolumn{1}{l}{Rec.}            & F1             \\ \hline
\multicolumn{6}{c}{Traditional Machine Learning}                                                                                                                                                                                         \\ \hline
\multicolumn{1}{l}{NB}                          & \multicolumn{1}{l}{BoW}                      & \multicolumn{1}{l}{76.38}                & \multicolumn{1}{l}{39.77}          & \multicolumn{1}{l}{43.21}          & 41.42          \\ \hline
\multicolumn{1}{l}{\multirow{3}{*}{SVM}}        & \multicolumn{1}{l}{TF-IDF}                   & \multicolumn{1}{l}{79.28}                & \multicolumn{1}{l}{49.21}          & \multicolumn{1}{l}{38.27}          & 43.06          \\  
\multicolumn{1}{l}{}                            & \multicolumn{1}{l}{Word2Vec}                 & \multicolumn{1}{l}{74.44}                & \multicolumn{1}{l}{39.20}          & \multicolumn{1}{l}{\textbf{60.49}} & 47.57          \\ 
\multicolumn{1}{l}{}                            & \multicolumn{1}{l}{IndoBERT}                 & \multicolumn{1}{l}{75.12}                & \multicolumn{1}{l}{35.65}          & \multicolumn{1}{l}{50.62}          & 41.84          \\ \hline
\multicolumn{1}{l}{\multirow{3}{*}{LR}}         & \multicolumn{1}{l}{TF-IDF}                   & \multicolumn{1}{l}{80.25}                & \multicolumn{1}{l}{53.19}          & \multicolumn{1}{l}{30.86}          & 39.06          \\ 
\multicolumn{1}{l}{}                            & \multicolumn{1}{l}{Word2Vec}                 & \multicolumn{1}{l}{73.86}                & \multicolumn{1}{l}{33.07}          & \multicolumn{1}{l}{51.85}          & 40.38          \\ 
\multicolumn{1}{l}{}                            & \multicolumn{1}{l}{IndoBERT}                 & \multicolumn{1}{l}{75.51}                & \multicolumn{1}{l}{35.54}          & \multicolumn{1}{l}{53.09}          & 42.57          \\ \hline
\multicolumn{1}{l}{\multirow{3}{*}{DT}}         & \multicolumn{1}{l}{TF-IDF}                   & \multicolumn{1}{l}{71.54}                & \multicolumn{1}{l}{30.23}          & \multicolumn{1}{l}{32.10}          & 31.14          \\ 
\multicolumn{1}{l}{}                            & \multicolumn{1}{l}{Word2Vec}                 & \multicolumn{1}{l}{59.63}                & \multicolumn{1}{l}{12.23}          & \multicolumn{1}{l}{20.99}          & 15.45          \\  
\multicolumn{1}{l}{}                            & \multicolumn{1}{l}{IndoBERT}                 & \multicolumn{1}{l}{64.67}                & \multicolumn{1}{l}{19.84}          & \multicolumn{1}{l}{30.86}          & 24.15          \\ \hline
\multicolumn{1}{l}{\multirow{3}{*}{RF}}         & \multicolumn{1}{l}{TF-IDF}                   & \multicolumn{1}{l}{80.35}                & \multicolumn{1}{l}{63.16}          & \multicolumn{1}{l}{29.63}          & 40.34          \\ 
\multicolumn{1}{l}{}                            & \multicolumn{1}{l}{Word2Vec}                 & \multicolumn{1}{l}{79.48}                & \multicolumn{1}{l}{51.56}          & \multicolumn{1}{l}{40.74}          & 45.52          \\ 
\multicolumn{1}{l}{}                            & \multicolumn{1}{l}{IndoBERT}                 & \multicolumn{1}{l}{78.51}                & \multicolumn{1}{l}{53.33}          & \multicolumn{1}{l}{39.51}          & 45.39          \\ \hline
\multicolumn{1}{l}{\multirow{3}{*}{XGB}}        & \multicolumn{1}{l}{TF-IDF}                   & \multicolumn{1}{l}{77.93}                & \multicolumn{1}{l}{\textbf{64.29}} & \multicolumn{1}{l}{22.22}          & 33.03          \\ 
\multicolumn{1}{l}{}                            & \multicolumn{1}{l}{Word2Vec}                 & \multicolumn{1}{l}{79.38}                & \multicolumn{1}{l}{47.19}          & \multicolumn{1}{l}{51.85}          & 49.41          \\  
\multicolumn{1}{l}{}                            & \multicolumn{1}{l}{IndoBERT}                 & \multicolumn{1}{l}{78.80}                & \multicolumn{1}{l}{44.19}          & \multicolumn{1}{l}{46.91}          & 45.51          \\ \hline
\multicolumn{6}{c}{Deep Learning}                                                                                                                                                                                                        \\ \hline
\multicolumn{1}{l}{BERT}                        & \multicolumn{1}{l}{IndoBERT}                 & \multicolumn{1}{l}{\textbf{85.00}}       & \multicolumn{1}{l}{55.68}          & \multicolumn{1}{l}{\textbf{60.49}} & \textbf{57.99} \\ 
\multicolumn{1}{l}{Bi-LSTM}                     & \multicolumn{1}{l}{IndoBERT}                 & \multicolumn{1}{l}{83.16}                & \multicolumn{1}{l}{57.63}          & \multicolumn{1}{l}{41.98}          & 48.57          \\ 
\multicolumn{1}{l}{DNN}                         & \multicolumn{1}{l}{IndoBERT}                 & \multicolumn{1}{l}{81.99}                & \multicolumn{1}{l}{56.39}          & \multicolumn{1}{l}{38.27}          & 45.59          \\ 
\multicolumn{1}{l}{CNN}                         & \multicolumn{1}{l}{IndoBERT}                 & \multicolumn{1}{l}{81.61}                & \multicolumn{1}{l}{57.14}          & \multicolumn{1}{l}{29.63}          & 39.02          \\ 
\hline
\multicolumn{6}{l}{\footnotesize{*Acc.: Accuracy; Prec.: Precision; Rec.: Recall; F1: F1-score}} 
\end{tabular}
\end{table}

\subsection{Two-Stage Classifier Results}

Next, we experimented with a pipeline model consisting of two sequential classifiers in which each was trained for different binary classification tasks. The first task was tweet's relevance classification and the second task was the misinformation detection.

\subsubsection{Relevance Classification Results}

The experimental results for the relevance classification task is shown on Table \ref{relevanceClassificationResults}. Based on that result, the IndoBERT model had the highest performance when compared to the other feature-classifier combinations. All of the performance metrics for this task were also led by IndoBERT. The difference between IndoBERT and the other models was relatively high. 


\begin{table}[t!]
\centering
\caption{Relevance Classification Results}
	\label{relevanceClassificationResults}
\begin{tabular}{llllll}
\multicolumn{1}{l}{\multirow{2}{*}{Classifier}} & \multicolumn{1}{l}{\multirow{2}{*}{Feature}} & \multicolumn{1}{l}{\multirow{2}{*}{Acc.}} & \multicolumn{3}{l}{Target = Relevance}                                                             \\ \cline{4-6} 
\multicolumn{1}{l}{}                            & \multicolumn{1}{l}{}                         & \multicolumn{1}{l}{}                     & \multicolumn{1}{l}{Prec.}           & \multicolumn{1}{l}{Rec.}            & F1             \\ \hline
\multicolumn{6}{c}{Traditional Machine Learning}                                                                                                                                                                                         \\ \hline
\multicolumn{1}{l}{NB}                          & \multicolumn{1}{l}{BoW}                      & \multicolumn{1}{l}{80.83}                & \multicolumn{1}{l}{74.13}          & \multicolumn{1}{l}{78.87}          & 76.43          \\ \hline
\multicolumn{1}{l}{\multirow{3}{*}{SVM}}        & \multicolumn{1}{l}{TF-IDF}                   & \multicolumn{1}{l}{83.16}                & \multicolumn{1}{l}{79.35}          & \multicolumn{1}{l}{77.40}          & 78.36          \\  
\multicolumn{1}{l}{}                            & \multicolumn{1}{l}{Word2Vec}                 & \multicolumn{1}{l}{80.45}                & \multicolumn{1}{l}{73.46}          & \multicolumn{1}{l}{78.87}          & 76.07          \\  
\multicolumn{1}{l}{}                            & \multicolumn{1}{l}{IndoBERT}                 & \multicolumn{1}{l}{79.48}                & \multicolumn{1}{l}{78.93}          & \multicolumn{1}{l}{65.36}          & 71.51          \\ \hline
\multicolumn{1}{l}{\multirow{3}{*}{LR}}         & \multicolumn{1}{l}{TF-IDF}                   & \multicolumn{1}{l}{83.54}                & \multicolumn{1}{l}{83.76}          & \multicolumn{1}{l}{72.24}          & 77.57          \\  
\multicolumn{1}{l}{}                            & \multicolumn{1}{l}{Word2Vec}                 & \multicolumn{1}{l}{81.12}                & \multicolumn{1}{l}{74.42}          & \multicolumn{1}{l}{79.36}          & 76.81          \\  
\multicolumn{1}{l}{}                            & \multicolumn{1}{l}{IndoBERT}                 & \multicolumn{1}{l}{81.61}                & \multicolumn{1}{l}{76.27}          & \multicolumn{1}{l}{77.40}          & 76.83          \\ \hline
\multicolumn{1}{l}{\multirow{3}{*}{DT}}         & \multicolumn{1}{l}{TF-IDF}                   & \multicolumn{1}{l}{78.99}                & \multicolumn{1}{l}{74.11}          & \multicolumn{1}{l}{71.74}          & 72.91          \\  
\multicolumn{1}{l}{}                            & \multicolumn{1}{l}{Word2Vec}                 & \multicolumn{1}{l}{71.93}                & \multicolumn{1}{l}{63.57}          & \multicolumn{1}{l}{67.32}          & 65.39          \\  
\multicolumn{1}{l}{}                            & \multicolumn{1}{l}{IndoBERT}                 & \multicolumn{1}{l}{70.76}                & \multicolumn{1}{l}{63.43}          & \multicolumn{1}{l}{60.93}          & 62.16          \\ \hline
\multicolumn{1}{l}{\multirow{3}{*}{RF}}         & \multicolumn{1}{l}{TF-IDF}                   & \multicolumn{1}{l}{83.64}                & \multicolumn{1}{l}{80.67}          & \multicolumn{1}{l}{76.90}          & 78.74          \\  
\multicolumn{1}{l}{}                            & \multicolumn{1}{l}{Word2Vec}                 & \multicolumn{1}{l}{82.28}                & \multicolumn{1}{l}{80.27}          & \multicolumn{1}{l}{72.97}          & 76.45          \\  
\multicolumn{1}{l}{}                            & \multicolumn{1}{l}{IndoBERT}                 & \multicolumn{1}{l}{80.93}                & \multicolumn{1}{l}{81.44}          & \multicolumn{1}{l}{66.83}          & 73.41          \\ \hline
\multicolumn{1}{l}{\multirow{3}{*}{XGB}}        & \multicolumn{1}{l}{TF-IDF}                   & \multicolumn{1}{l}{79.38}                & \multicolumn{1}{l}{83.45}          & \multicolumn{1}{l}{59.46}          & 69.44          \\  
\multicolumn{1}{l}{}                            & \multicolumn{1}{l}{Word2Vec}                 & \multicolumn{1}{l}{83.83}                & \multicolumn{1}{l}{78.30}          & \multicolumn{1}{l}{81.57}          & 79.90          \\  
\multicolumn{1}{l}{}                            & \multicolumn{1}{l}{IndoBERT}                 & \multicolumn{1}{l}{81.90}                & \multicolumn{1}{l}{77.09}          & \multicolumn{1}{l}{76.90}          & 77.00          \\ \hline
\multicolumn{6}{c}{Deep Learning}                                                                                                                                                                                                        \\ \hline
\multicolumn{1}{l}{BERT}                        & \multicolumn{1}{l}{IndoBERT}                 & \multicolumn{1}{l}{\textbf{90.03}}       & \multicolumn{1}{l}{\textbf{87.62}} & \multicolumn{1}{l}{\textbf{86.98}} & \textbf{87.30} \\ 
\multicolumn{1}{l}{Bi-LSTM}                     & \multicolumn{1}{l}{IndoBERT}                 & \multicolumn{1}{l}{85.09}                & \multicolumn{1}{l}{79.91}          & \multicolumn{1}{l}{83.05}          & 81.45          \\ 
\multicolumn{1}{l}{DNN}                         & \multicolumn{1}{l}{IndoBERT}                 & \multicolumn{1}{l}{84.22}                & \multicolumn{1}{l}{78.50}          & \multicolumn{1}{l}{82.56}          & 80.48          \\ 
\multicolumn{1}{l}{CNN}                         & \multicolumn{1}{l}{IndoBERT}                 & \multicolumn{1}{l}{83.45}                & \multicolumn{1}{l}{78.23}          & \multicolumn{1}{l}{80.34}          & 79.27          \\ \hline
\multicolumn{6}{l}{\footnotesize{*Acc.: Accuracy; Prec.: Precision; Rec.: Recall; F1: F1-score}} 
\end{tabular}
\end{table}

\subsubsection{Misinformation Classification Results}

For this task, only the tweets with true or misinformation labels were included in the training, validation, and testing process. The results  are shown in Table \ref{misinformationClassificationResults}. We found that IndoBERT did not perform as well as on the previous tasks. Instead, the Bi-LSTM with IndoBERT gained the highest accuracy, precision, and F1 scores. Other combinations that performed nearly as well as the best one were XGBoost-IndoBERT, Logistic Regression-IndoBERT, Random Forest-IndoBERT, and CNN-IndoBERT. The top five combinations were using IndoBERT embeddings as their feature input. 


\begin{table}[t!]
\centering
\caption{Misinformation Classification Results}
	\label{misinformationClassificationResults}
\begin{tabular}{llllll}
\multicolumn{1}{l}{\multirow{2}{*}{Classifier}} & \multicolumn{1}{l}{\multirow{2}{*}{Feature}} & \multicolumn{1}{l}{\multirow{2}{*}{Acc.}} & \multicolumn{3}{l}{Target = Misinformation}                                                             \\ \cline{4-6} 
\multicolumn{1}{l}{}                            & \multicolumn{1}{l}{}                         & \multicolumn{1}{l}{}                     & \multicolumn{1}{l}{Prec.}           & \multicolumn{1}{l}{Rec.}            & F1             \\ \hline
\multicolumn{6}{c}{Traditional Machine Learning}                                                                                                                                                                                         \\ \hline
\multicolumn{1}{l}{NB}                          & \multicolumn{1}{l}{BoW}                      & \multicolumn{1}{l}{86.98}                & \multicolumn{1}{l}{66.67}          & \multicolumn{1}{l}{69.14}          & 67.88          \\ \hline
\multicolumn{1}{l}{\multirow{3}{*}{SVM}}        & \multicolumn{1}{l}{TF-IDF}                   & \multicolumn{1}{l}{88.21}                & \multicolumn{1}{l}{71.43}          & \multicolumn{1}{l}{67.90}          & 69.62          \\  
\multicolumn{1}{l}{}                            & \multicolumn{1}{l}{Word2Vec}                 & \multicolumn{1}{l}{84.28}                & \multicolumn{1}{l}{58.25}          & \multicolumn{1}{l}{74.07}          & 65.22          \\  
\multicolumn{1}{l}{}                            & \multicolumn{1}{l}{IndoBERT}                 & \multicolumn{1}{l}{84.52}                & \multicolumn{1}{l}{59.00}          & \multicolumn{1}{l}{72.84}          & 65.19          \\ \hline
\multicolumn{1}{l}{\multirow{3}{*}{LR}}         & \multicolumn{1}{l}{TF-IDF}                   & \multicolumn{1}{l}{88.94}                & \multicolumn{1}{l}{\textbf{80.00}} & \multicolumn{1}{l}{59.26}          & 68.09          \\  
\multicolumn{1}{l}{}                            & \multicolumn{1}{l}{Word2Vec}                 & \multicolumn{1}{l}{86.98}                & \multicolumn{1}{l}{64.89}          & \multicolumn{1}{l}{75.31}          & 69.71          \\  
\multicolumn{1}{l}{}                            & \multicolumn{1}{l}{IndoBERT}                 & \multicolumn{1}{l}{89.43}                & \multicolumn{1}{l}{71.59}          & \multicolumn{1}{l}{77.78}          & 74.56          \\ \hline
\multicolumn{1}{l}{\multirow{3}{*}{DT}}         & \multicolumn{1}{l}{TF-IDF}                   & \multicolumn{1}{l}{80.84}                & \multicolumn{1}{l}{51.65}          & \multicolumn{1}{l}{58.02}          & 54.65          \\  
\multicolumn{1}{l}{}                            & \multicolumn{1}{l}{Word2Vec}                 & \multicolumn{1}{l}{79.85}                & \multicolumn{1}{l}{49.53}          & \multicolumn{1}{l}{65.43}          & 56.38          \\  
\multicolumn{1}{l}{}                            & \multicolumn{1}{l}{IndoBERT}                 & \multicolumn{1}{l}{77.64}                & \multicolumn{1}{l}{46.03}          & \multicolumn{1}{l}{71.60}          & 56.04          \\ \hline
\multicolumn{1}{l}{\multirow{3}{*}{RF}}         & \multicolumn{1}{l}{TF-IDF}                   & \multicolumn{1}{l}{85.75}                & \multicolumn{1}{l}{68.25}          & \multicolumn{1}{l}{53.09}          & 59.72          \\  
\multicolumn{1}{l}{}                            & \multicolumn{1}{l}{Word2Vec}                 & \multicolumn{1}{l}{88.21}                & \multicolumn{1}{l}{70.37}          & \multicolumn{1}{l}{70.37}          & 70.37          \\  
\multicolumn{1}{l}{}                            & \multicolumn{1}{l}{IndoBERT}                 & \multicolumn{1}{l}{89.93}                & \multicolumn{1}{l}{72.73}          & \multicolumn{1}{l}{\textbf{79.01}} & 75.74          \\ \hline
\multicolumn{1}{l}{\multirow{3}{*}{XGB}}        & \multicolumn{1}{l}{TF-IDF}                   & \multicolumn{1}{l}{84.28}                & \multicolumn{1}{l}{64.91}          & \multicolumn{1}{l}{45.68}          & 53.62          \\  
\multicolumn{1}{l}{}                            & \multicolumn{1}{l}{Word2Vec}                 & \multicolumn{1}{l}{86.98}                & \multicolumn{1}{l}{65.22}          & \multicolumn{1}{l}{74.07}          & 69.36          \\  
\multicolumn{1}{l}{}                            & \multicolumn{1}{l}{IndoBERT}                 & \multicolumn{1}{l}{90.42}                & \multicolumn{1}{l}{74.42}          & \multicolumn{1}{l}{\textbf{79.01}} & 76.65          \\ \hline
\multicolumn{6}{c}{Deep Learning}                                                                                                                                                                                                        \\ \hline
\multicolumn{1}{l}{BERT}                        & \multicolumn{1}{l}{IndoBERT}                 & \multicolumn{1}{l}{88.48}                & \multicolumn{1}{l}{74.29}          & \multicolumn{1}{l}{64.20}          & 68.87          \\ 
\multicolumn{1}{l}{Bi-LSTM}                     & \multicolumn{1}{l}{IndoBERT}                 & \multicolumn{1}{l}{\textbf{91.15}}       & \multicolumn{1}{l}{\textbf{80.00}} & \multicolumn{1}{l}{74.07}          & \textbf{76.92} \\ 
\multicolumn{1}{l}{DNN}                         & \multicolumn{1}{l}{IndoBERT}                 & \multicolumn{1}{l}{89.93}                & \multicolumn{1}{l}{78.57}          & \multicolumn{1}{l}{67.90}          & 72.85          \\ 
\multicolumn{1}{l}{CNN}                         & \multicolumn{1}{l}{IndoBERT}                 & \multicolumn{1}{l}{90.17}                & \multicolumn{1}{l}{75.95}          & \multicolumn{1}{l}{74.07}          & 75.00          \\ \hline
\multicolumn{6}{l}{\footnotesize{*Acc.: Accuracy; Prec.: Precision; Rec.: Recall; F1: F1-score}} 
\end{tabular}
\end{table}
\begin{table}[t!]
\centering
\caption{Pipeline Results}
	\label{pipelineResults}
\begin{tabular}{lllllll}
\multicolumn{2}{c}{Classifier}                                                  & \multicolumn{1}{l}{\multirow{2}{*}{Feature}} & \multicolumn{1}{l}{\multirow{2}{*}{Acc.}} & \multicolumn{3}{l}{Target = Misinformation}                                                        \\ \cline{1-2} \cline{5-7} 
\multicolumn{1}{l}{Relev.}             & \multicolumn{1}{l}{Misinfo.} & \multicolumn{1}{l}{}                         & \multicolumn{1}{l}{}                     & \multicolumn{1}{l}{Prec.}           & \multicolumn{1}{l}{Rec.}            & F1             \\ \hline
\multicolumn{7}{c}{Single Classifier}                                                                                                                                                                                                                                    \\ \hline
\multicolumn{2}{c}{BERT}                                                        & \multicolumn{1}{l}{IndoBERT}                 & \multicolumn{1}{l}{85.00}                & \multicolumn{1}{l}{55.68}          & \multicolumn{1}{l}{60.49}          & 57.99          \\ 
\multicolumn{2}{c}{XGB}                                                         & \multicolumn{1}{l}{Word2Vec}                 & \multicolumn{1}{l}{79.38}                & \multicolumn{1}{l}{47.19}          & \multicolumn{1}{l}{51.85}          & 49.41          \\ 
\multicolumn{2}{c}{Bi-LSTM}                                                     & \multicolumn{1}{l}{IndoBERT}                 & \multicolumn{1}{l}{83.16}                & \multicolumn{1}{l}{57.63}          & \multicolumn{1}{l}{41.98}          & 48.57          \\ 
\multicolumn{2}{c}{SVM}                                                         & \multicolumn{1}{l}{Word2Vec}                 & \multicolumn{1}{l}{74.44}                & \multicolumn{1}{l}{39.20}          & \multicolumn{1}{l}{60.49}          & 47.57          \\ 
\multicolumn{2}{c}{DNN}                                                         & \multicolumn{1}{l}{IndoBERT}                 & \multicolumn{1}{l}{81.99}                & \multicolumn{1}{l}{56.36}          & \multicolumn{1}{l}{38.27}          & 45.59          \\ \hline
\multicolumn{7}{c}{Two-Stage Classifier Pipeline}                                                                                                                                                                                                                            \\ \hline
\multicolumn{1}{l}{\multirow{5}{*}{BERT}} & \multicolumn{1}{l}{Bi-LSTM}        & \multicolumn{1}{l}{IndoBERT}                 & \multicolumn{1}{l}{\textbf{87.02}}       & \multicolumn{1}{l}{\textbf{59.75}} & \multicolumn{1}{l}{60.49}          & \textbf{60.12} \\ 
\multicolumn{1}{l}{}                      & \multicolumn{1}{l}{XGB}            & \multicolumn{1}{l}{IndoBERT}                 & \multicolumn{1}{l}{85.77}                & \multicolumn{1}{l}{55.43}          & \multicolumn{1}{l}{62.96}          & 58.96          \\  
\multicolumn{1}{l}{}                      & \multicolumn{1}{l}{LR}             & \multicolumn{1}{l}{IndoBERT}                 & \multicolumn{1}{l}{85.58}                & \multicolumn{1}{l}{54.17}          & \multicolumn{1}{l}{\textbf{64.20}} & 58.76          \\ 
\multicolumn{1}{l}{}                      & \multicolumn{1}{l}{RF}             & \multicolumn{1}{l}{IndoBERT}                 & \multicolumn{1}{l}{85.77}                & \multicolumn{1}{l}{54.84}          & \multicolumn{1}{l}{62.96}          & 58.62          \\ 
\multicolumn{1}{l}{}                      & \multicolumn{1}{l}{CNN}            & \multicolumn{1}{l}{IndoBERT}                 & \multicolumn{1}{l}{86.74}                & \multicolumn{1}{l}{54.35}          & \multicolumn{1}{l}{61.73}          & 57.80          \\ \hline
\multicolumn{7}{l}{\footnotesize{*Relev.: Relevance; Misinfo.: Misinformation}} \\ 
\multicolumn{7}{l}{\footnotesize{**Acc.: Accuracy; Prec.: Precision; Rec.: Recall; F1: F1-score}} 
\end{tabular}
\end{table}

\subsubsection{Pipeline Comparison}

We compared our proposed two-stage classifier pipeline with the baseline single 3-class classifier. For each method, five of the best combinations were picked to be compared. Based on the results in Table \ref{pipelineResults}, the best outcome was achieved by combining BERT for relevance classification and Bi-LSTM for misinformation classification using IndoBERT as its feature with an accuracy of 87.02\% and F1 of 60.12\%.

Apart from BERT+CNN, we found that the other top four models on the two-stage Classifier pipeline achieved higher scores compared to the single classifier model. Our proposed method by splitting the specific tasks yielded substantially better results compared to using a multiclass classification. Other than using the two-stage classifier pipeline, it was also indicated that using pre-trained embeddings (Word2Vec \cite{karyana_2018} and IndoBERT \cite{wilie2020indonlu}) produced better results when compared to using more traditional features such as BoW and TF-IDF.

\subsection{Error Analysis}


\begin{figure}[t!]
     \centering
     \begin{subfigure}
         \centering
         \includegraphics[width=0.5\textwidth]{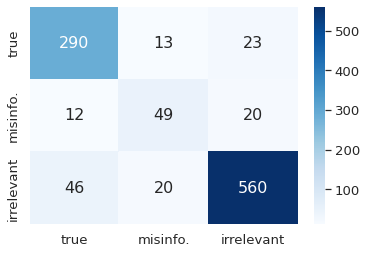}
         \caption{Ternary Class Confusion Matrix from The Best Model.}
         \label{figConfMatrix3}
     \end{subfigure}
     \begin{subfigure}
         \centering
         \includegraphics[width=0.5\textwidth]{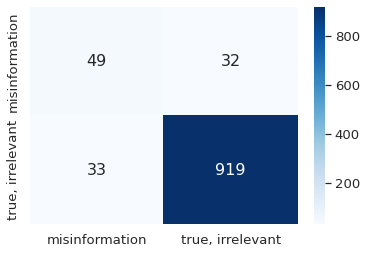}
         \caption{Binary Class Confusion Matrix from The Best Model.}
         \label{figConfMatrixBinary}
     \end{subfigure}
\end{figure}

\begin{table}[t!]
    \centering
    \begin{tabular}{llp{90pt}p{90pt}}
        Annotation     & Predicted      & Tweet Text                                                                                                                               & Translated                                                                                                                                         \\ \hline
Irrelevant     & Misinformation & \textit{Anies Bersyukur DKI Pecah Rekor Positif Corona, ini hasil Kita kerja Senyap dan Sunyi,,, kerja Senyap dan Sunyi kok anggaran juga Amblas} & "Anies is grateful that DKI has a positive record of Corona, this is the result by persevering quietly"... If so, explain why the budgeting failed \\ 
Misinformation & Irrelevant     & \textit{VAKSIN CORONA !!! SUDAH ADA YANG. HALAL !!! KENAPA MAKSA !!!YANG VAKSIN HARAM !!! JELAS SEKALI KEJAHATANNYA !!!}                          & CORONA VACCINE !!! THERE IS THE HALAL OPTION !!! WHY ARE WE FORCED WITH THE NON-HALAL !!! OBVIOUSLY A CRIME !!!                                    \\ 
Misinformation & True           & \textit{RSUD Banten Karantina 7 Mahasiwa Dari Wuhan Soal Virus Corona}                                                                            & The Public Hospital in Banten Quarantined 7 Students from Wuhan Due to Coronavirus precaution                                                                \\ 
    \end{tabular}
    \caption{Samples of False Predictions}
    \label{falsePreds}
\end{table}


We took the best model, which was BERT+Bi-LSTM to analyze the output predictions. The confusion matrix for the three classes and two classes classification are shown in Figure \ref{figConfMatrix3} and Figure \ref{figConfMatrixBinary} respectively. 

The imbalanced nature of the dataset had an impact on the classifiers especially in classifying the misinformation class. Both the true and irrelevant classes were relatively easy to predict by the model. On the other hand, the misinformation class was relatively more difficult to predict. 

If the performance is to be viewed solely as a misinformation detection performance, as shown in Figure \ref{figConfMatrixBinary}, the model still performed well despite a highly imbalanced dataset. Based on the numbers of false positives and false negatives, the model also managed to find a good balance between precision and recall, which resulted in almost no performance trade-off in either metric (precision and recall).

Several false positives and false negatives were sampled to provide a case-by-case error analysis shown in Table \ref{falsePreds}. In the first case, the tweet author expressed their disapproval towards The Governor of Jakarta Capital Region, Anies Baswedan, who had stated that Jakarta was showing improvements in controlling the spread of the coronavirus thanks to the government's effort. During the annotation process, we decided to mark any politically-related tweet as irrelevant. However, the model predicted the tweet as a misinformation tweet.

Both the latter cases were false negatives. For the second case, it should be noted that Indonesia is majorly populated by Muslims. Their belief (Islam) does not allow the consumption or any means of inserting non-\textit{halal} or \textit{haram} substances such as pork or dog meat into one's body. During the beginning of 2021 when COVID-19 vaccines were starting to be distributed, there were uncertainties regarding whether the vaccines were made out of \textit{halal} materials. The tweet author claimed that the vaccines distributed by the government at that time, which was CoronaVac by Sinovac, was not \textit{halal}, and blamed the government for forcing the Muslims to get vaccinated. However, in this case, the model predicted this as irrelevant and failed to flag this as misinformation.

The third case was from a news outlet that had shared a headline on Twitter. It turns out that there was a revision to the headline in the actual article itself. Before the revision, the headline stated that 7 students were quarantined. It was found later that there were 9 students instead of 7. The annotators annotated this as misinformation due to the discrepancy between the tweet and the revised article. However, the model predicted this as true. The most possible explanation is that the text's semantics were similar to other headlines which were labeled as true.



\section{Conclusion}

It is difficult to predict when the spread of the COVID-19 would ease up and no longer be declared a pandemic. False claims about the disease would continue to be made and spread through different types of mediums. Moving forward, the need to detect misinformation spread on social media would remain not only around COVID-19 but also for different topics.

In this research, we have constructed dataset and conducted the experiments for detecting COVID-19 misinformation that is spreading through social media. Previous works were studied and synthesized to design a robust dataset following a thorough annotation guideline. 
We formulated the misinformation detection task as two serialized subtasks, i.e., relevance classification and misinformation classification. 
Splitting the different tasks and using a pre-trained language model improved the overall prediction performance.

For future research, the misinformation detection task may incorporate language models and information retrieval techniques to automatically verify the factual claims that are contained in the social media posts. To broaden the research scope, the inclusion of other social media platforms can also be considered. Lastly, the lack of annotated data could become an issue, especially in low-resource languages, and approaches such as semi-supervised learning can be explored for this task.

\appendix

\section{Annotation Guideline}

The following guidelines were shared with the annotators involved in this research. The guidelines were originally written in Indonesian and then translated into English by the authors. 


\begin{figure}
    \centering
    \includegraphics[width=\textwidth]{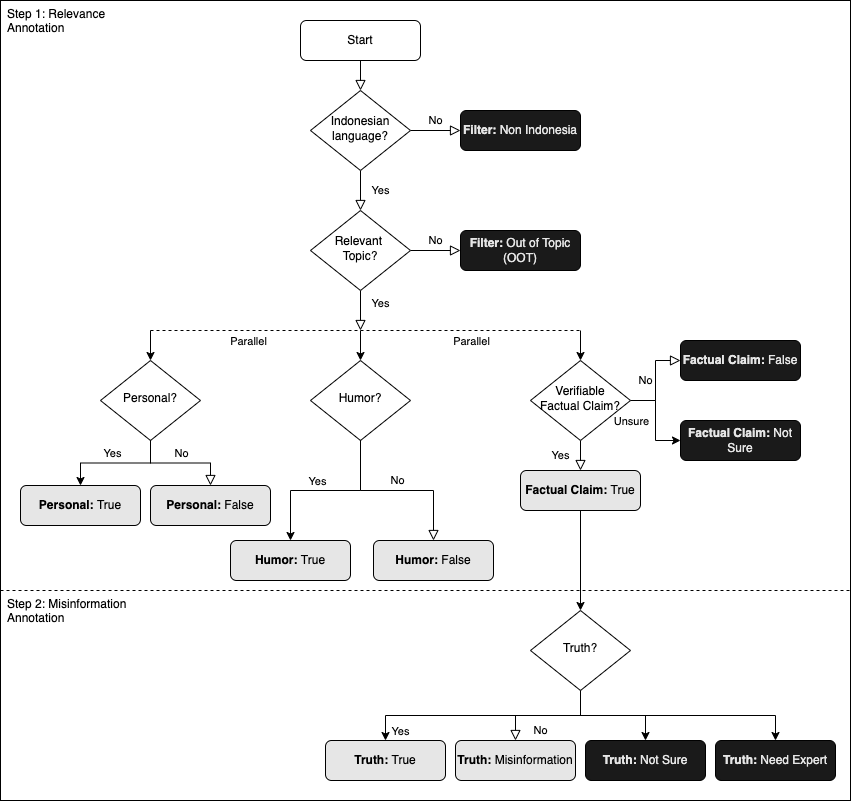}
    \caption{Annotation Diagram.}
    \label{figAnnotationDiagram}
\end{figure}

\subsection{Relevance Annotation Guideline}

\subsubsection{Task Definition}

Annotations are to be made on Indonesian tweet data with topics revolving around COVID-19. The expected output is a collection of labeled tweets that have been annotated based on topic relevance and whether the tweet contains \textit{verifiable factual claim}.

The term \textit{verifiable factual claim} refers to the definition made by \cite{alam2021fighting} in their annotation guideline. A verifiable factual claim is a statement that claims something or some events are true and its truth can be verified using any publicly available factual verified information such as statistics, specific examples, or personal testimonials. The factual claim itself includes:
\begin{itemize}
\item A statement
\item Any mentions of a quantity value in the present or the past
\item Any predictions of the future that is verifiable
\item A reference to a product of law, procedures, or operational rules
\item A correlation or causal statement
\end{itemize}

\subsubsection{Data Columns}

\begin{enumerate}
\item Tweet Columns
\begin{itemize}
\item \texttt{tweet\_url}: Anonymized link to the related tweet
\item \texttt{text}: The text in the tweet post
\item \texttt{urls}: Links that are written in the tweet
\item \texttt{date}: The date when the tweet was posted
\end{itemize}
\item Annotation Columns 
\begin{itemize}
\item \texttt{filter}: Determine the initial relevance of the tweet (relevant / non-Indonesian / out-of-topic)
\item \texttt{personal}: Whether the tweet contains something personal \cite{dann2010twitter,yen2018detecting} (true/false)
\item \texttt{humor}: Whether the tweet written for humor purposes \cite{zhang2014recognizing} (true/false)
\item \texttt{factual\_claim}: Whether the tweet fulfills the criteria of a verifiable factual claim \cite{alam2021fighting} (true / not sure / false)
\end{itemize}
\end{enumerate}

\subsubsection{Annotation Steps}

\begin{enumerate}
\item Identify the languages used in the tweet.
\begin{enumerate}
\item If at least one of the following criteria is fulfilled, proceed to step \ref{relAnnotStep2}.
\begin{itemize}
\item The entire tweet is written in Indonesian.
\item The Tweet is code-mixed, e.g., Indonesian-English \cite{barik2019normalization} with the majority text content is written in Indonesian.
\end{itemize}
\item If the criteria is not fulfilled, annotate the column \texttt{filter} as \textbf{non-indonesia}.
\end{enumerate}
\item \label{relAnnotStep2} Identify the topic discussed in the tweet.
\begin{enumerate}
\item If the tweet discussed any of the following topics, annotate the column \texttt{filter} as \textbf{out-of-topic}.
\begin{itemize}
\item Political campaign or critic.
\item Discussed a certain topic but used a highly unrelated yet trending hashtag.
\item Any forms of advertisements.
\end{itemize}
\item If the main topic discussed in the tweet involved one of the following topics, annotate the column \texttt{filter} as \textbf{relevant} and proceed to step \ref{relAnnotStep3}.
\begin{itemize}
\item The COVID-19 pandemic situation in Indonesia or globally.
\item Scientific discoveries, vaccines, medicines, or medical treatments related to COVID-19.
\item Any official policies related to handling the COVID-19 pandemic in Indonesia or globally.
\item Other affected sectors by the pandemic (socioeconomic for example)
\end{itemize}
\item If it does not fulfill the previous two criterias, annotate the columns \texttt{filter} as \textbf{out-of-topic}. 
\end{enumerate}
\item \label{relAnnotStep3} Identify Verifiable Factual Claim \cite{alam2021fighting} on the three columns as follow:
\begin{enumerate}
\item \texttt{personal}. If the tweet discussed a personal experience, annotate this column as \textbf{true}.
\item \texttt{humor}. If the tweet has a humorous intent, annotate this column as \textbf{true}.
\item \texttt{factual\_claim}. If the tweet contains a factual claim that can be verified with publicly available information, annotate this column as \textbf{true}. If you are unsure, annotate this column as \textbf{not sure}. Otherwise, annotate \textbf{false}.
\end{enumerate}
\end{enumerate}

\subsection{Misinformation Annotation Guideline}

\subsubsection{Task Definition}

Annotations are to be made on Indonesian tweet data with topics revolving around COVID-19. The expected output is a collection of tweets that have been annotated based on the truth or misinformation of the \textit{verifiable factual claim} contained within the tweets. The tweets that are given are already filtered based on their relevance. It should be noted that the annotation should only address the claim mentioned in the text. The other form of media attached to the tweet should not affect your judgment during the annotation process.

The term \textit{verifiable factual claim} refers to the definition made by \cite{alam2021fighting} in their annotation guideline. A verifiable factual claim is a statement that claims something or some events are true and its truth can be verified using any publicly available factual verified information such as statistics, specific examples, or personal testimonials. The factual claim itself includes:
\begin{itemize}
\item A statement
\item Any mentions of a quantity value in the present or the past
\item Any predictions on the future that is verifiable
\item A reference to a product of law, procedures, or operational rules
\item A correlation or causal statement
\end{itemize}

In this context, the misinformation in question is a factual claim that is:
\begin{itemize}
\item Partly or completely false. Example: "CAREFUL!! MICROCHIPS are planted within the VACCINES!".
\item Contains misleading or \textit{clickbait} elements. Example: "MUI (Indonesian Ulema Council) declared COVID vaccines as \textit{haram}!?".
\item Rumors or conspiracy theories that have not been proven. Example: "COVID will end in 2022".
\end{itemize}

\subsubsection{Data Columns}

\begin{enumerate}
\item Tweet Columns
\begin{itemize}
\item \texttt{tweet\_url}: Anonymized link to the related tweet
\item \texttt{text}: The text in the tweet post
\item \texttt{urls}: Links that are written in the tweet
\item \texttt{date}: The date when the tweet was posted
\end{itemize}
\item Annotation Columns 
\begin{itemize}
\item \texttt{truth}: The truth of the tweet (true / misinformation / not sure / need expert)
\end{itemize}
\end{enumerate}

\subsubsection{Steps}

\begin{enumerate}
    \item Verification Steps
    
    Tweet example: "It turned out that Chloroquine has been used to treat several COVID-19 patients in Indonesia" (March 23, 2020).
    
    \begin{enumerate}
        \item Determine the information that needs to be verified. Example based on the tweet above: "Indonesia chloroquine COVID treatment".
        \item Look for the information using a trusted search engine.
        \item If the tweet information matches the source of the article that can be considered reliable (reputable media or medical institutions), then the information can be considered true. Otherwise, the tweet may contain misinformation.
        \item You can open the external hyperlinks (\texttt{tweet\_url} or \texttt{urls}) to get more context of the tweet. However, please note that only the text content is considered for the annotation.
    \end{enumerate}
    
    \item Annotation Steps
    
    \begin{enumerate}
        \item If the information is true, annotate as \textbf{true}.
        \item If the information is false, partly false, or misleading, annotate as \textbf{misinformation}.
        \item If you are not sure about the truth of the claim, annotate as \textbf{not sure}.
        \item If you think an expert's annotation is needed for this tweet, annotate as \textbf{need expert}.
    \end{enumerate}

\end{enumerate}

\bibliographystyle{elsarticle-num} 
\bibliography{references}





\end{document}